\documentclass[letterpaper]{article} 
\usepackage[]{aaai24}  
\usepackage{times}  
\usepackage{helvet}  
\usepackage{courier}  
\usepackage[hyphens]{url}  
\usepackage{graphicx} 
\usepackage{natbib}  
\usepackage{caption} 
\usepackage{algorithm}
\usepackage{algorithmic}
\usepackage{xspace}
\usepackage{multirow}
\usepackage{array}
\usepackage{tabularx}
\usepackage[table,xcdraw]{xcolor}
\usepackage{booktabs} 
\usepackage{subcaption}
\usepackage{amsfonts} 
\usepackage{amsmath}
\usepackage{bm}

\usepackage{xcolor}
\usepackage{amsmath}
\usepackage{tabto}
\usepackage{multirow}
\usepackage{microtype}
\usepackage{booktabs}
\usepackage{enumitem}
\usepackage[utf8]{inputenc}
\usepackage{listings}
\usepackage{caption}
\usepackage{dsfont}
\usepackage{graphicx}
\usepackage{amsthm}
\usepackage{afterpage}
\usepackage{xspace}
\usepackage{adjustbox}
\usepackage{booktabs, tabularx}
\usepackage{makecell}
\usepackage{nicematrix,tikz}
\usepackage{color}
\usepackage{babel}
\usepackage{nameref}

\newcommand{\frameworkname}{Graph Neural Prompting\xspace}
\newcommand{\frameworkabbr}{GNP\xspace}

\usepackage{nicematrix,tikz}

\newtheorem{definition}{Definition}

\newtheorem{problem}{Problem}

\usepackage{newfloat}
\usepackage{listings}
\DeclareCaptionStyle{ruled}{labelfont=normalfont,labelsep=colon,strut=off} 
\floatstyle{ruled}
\newfloat{listing}{tb}{lst}{}
\floatname{listing}{Listing}

\title{Graph Neural Prompting with Large Language Models}

\author {
    Yijun Tian\textsuperscript{\rm 1},
    Huan Song\textsuperscript{\rm 2},
    Zichen Wang\textsuperscript{\rm 2},
    Haozhu Wang\textsuperscript{\rm 2},\\
    Ziqing Hu\textsuperscript{\rm 2},
    Fang Wang\textsuperscript{\rm 2},
    Nitesh V. Chawla\textsuperscript{\rm 1},
    Panpan Xu\textsuperscript{\rm 2}
}
\affiliations {
    \textsuperscript{\rm 1}University of Notre Dame\\
    \textsuperscript{\rm 2}Amazon\\
    \{yijun.tian, nchawla\}@nd.edu, 
    \{huanso, zichewan, haozhuw, ziqinghu, fwfang, xupanpan\}@amazon.com
}

\begin{document}

\maketitle

\begin{abstract}

Large language models (LLMs) have shown remarkable generalization capability with exceptional performance in various language modeling tasks. However, they still exhibit inherent limitations in precisely capturing and returning grounded knowledge. While existing work has explored utilizing knowledge graphs (KGs) to enhance language modeling via joint training and customized model architectures, applying this to LLMs is problematic owing to their large number of parameters and high computational cost. Therefore, how to enhance pre-trained LLMs using grounded knowledge, e.g., retrieval-augmented generation, remains an open question. In this work, we propose Graph Neural Prompting (GNP), a novel plug-and-play method to assist pre-trained LLMs in learning beneficial knowledge from KGs. GNP encompasses various designs, including a standard graph neural network encoder, a cross-modality pooling module, a domain projector, and a self-supervised link prediction objective. Extensive experiments on multiple datasets demonstrate the superiority of GNP on both commonsense and biomedical reasoning tasks across different LLM sizes and settings. Code is available at \mbox{\url{https://github.com/meettyj/GNP}}.

 
\end{abstract}

\section{Introduction}

Large Language Models (LLMs) have demonstrated exceptional performance and general capability in various NLP tasks and use cases such as question answering \cite{robinson2022leveraging} and text summarization \cite{zhang2023benchmarking}. Moreover, the significant growth in model size has further endowed LLMs with emergent capabilities \cite{wei2022emergent}, laying the groundwork for exploring artificial general intelligence \cite{bubeck2023sparks}. Accordingly, LLMs have attracted tremendous interest from academia \cite{wei2021finetuned, llm_survey} and industry \cite{anil2023palm, openai2023gpt4}.

Given the broad success of LLMs, many techniques have emerged to adapt these general-purpose models to downstream tasks. Beyond the conventional approach of model fine-tuning where all model parameters are adjusted \cite{howard2018universal}, prompt-based adaptation methods are proposed to modulate a frozen LLM's behavior through prompts \cite{prompt_design, prompt_tuning, prefix_tuning}. Rather than adapt the parameters in LLMs, these methods freeze the LLMs and typically introduce additional trainable parameters. The idea of freezing LLMs is appealing, especially as the model size grows and the training resource dependency intensifies.

\begin{figure}[t]
\centering
\includegraphics[width=\linewidth]{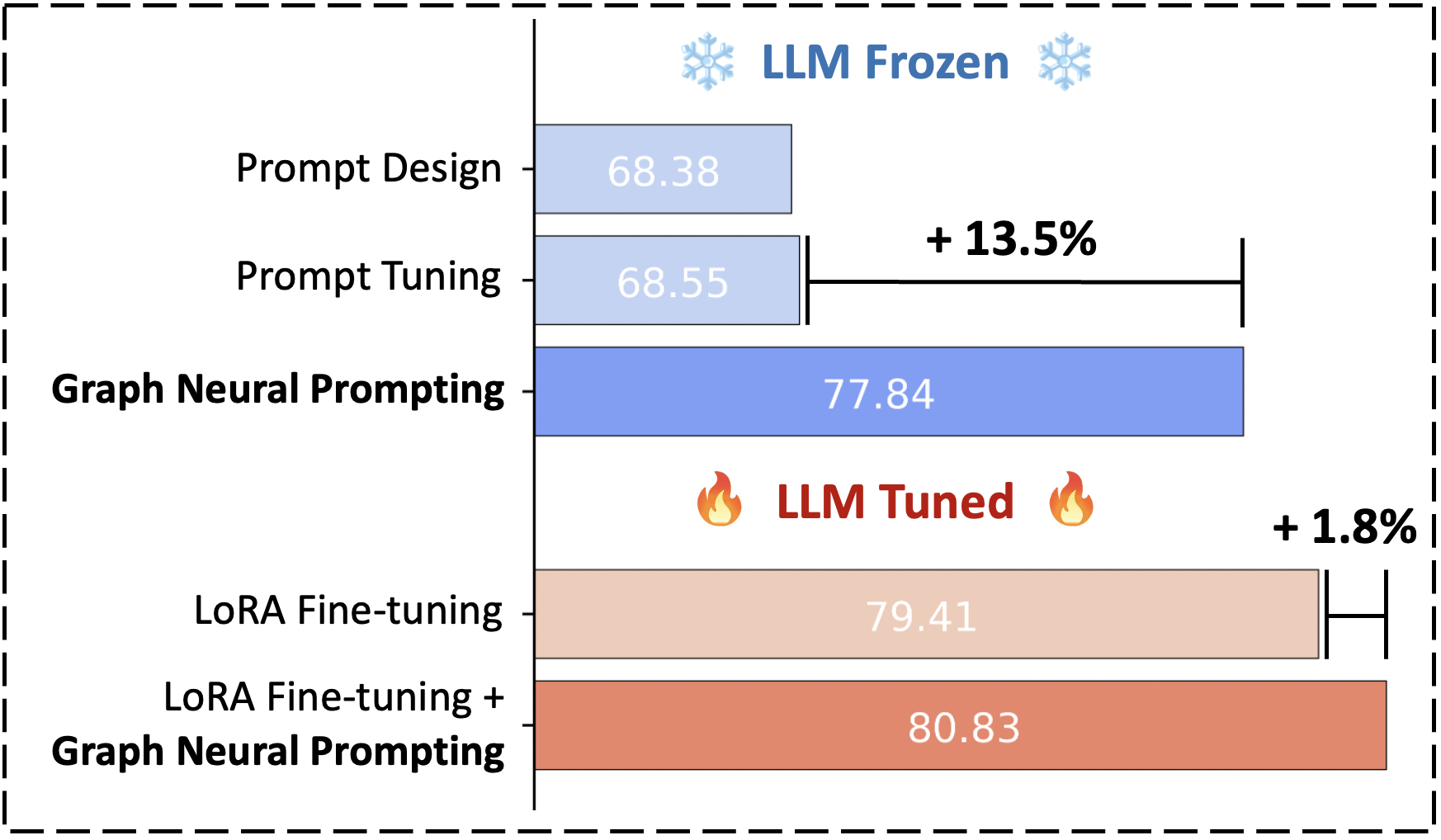}
\caption{
Result comparison across LLM Frozen (parameters unchanged) and LLM Tuned (parameters updated) settings. The proposed Graph Neural Prompting significantly improves the performance. Reported results are averaged across six datasets on two tasks for an 11B FLAN-T5 model.
}
\label{fig:intro_results}
\end{figure}

On the other hand, despite the success of LLMs in handling different real-world applications and the feasibility of adapting to specific downstream tasks, they still exhibit the inherent limitations of language modeling in accurately capturing and returning grounded knowledge \cite{lewis2020retrieval, kg_llm_survey}. Knowledge graphs (KGs), storing enormous facts, serve as a systematic way of representing knowledge \cite{kg_survey}. Consequently, existing methods have incorporated KGs to assist language modeling, often by designing customized model architectures to accommodate both KGs and textual data, followed by joint training sessions \cite{dragon, greaselm}. Nonetheless, joint training KGs and text for LLMs is challenging due to the extensive parameters LLMs contain and the substantial computation resources they require. In addition, numerous pre-trained LLMs with exceptional capabilities are released. It becomes advantageous to employ these pre-existing LLMs, particularly beneficial if we can sidestep the need to craft a specialized model and train it from scratch. A direct approach to employing KGs for retrieval-augmented generation \cite{lewis2020retrieval} is to feed the KG triples into LLMs directly \cite{kaping}. However, this method can introduce substantial noise, given that KGs might contain various extraneous contexts. Therefore, we ask: 
\begin{center}
\textbf{\textit{Can we learn beneficial knowledge from KGs \\ and integrate them into pre-trained LLMs?}}
\end{center}

\noindent
To answer the question, we propose \frameworkname (\frameworkabbr), a novel plug-and-play method to assist pre-trained LLMs in learning beneficial knowledge from KGs. GNP retrieves and encodes the pertinent grounded knowledge to derive Graph Neural Prompt, an embedding vector that can be sent into LLMs to provide guidance and instructions. In particular, GNP first utilizes a graph neural network (GNN) to capture and encode the intricate graph knowledge into entity/node embeddings. Then, a cross-modality pooling module is present to determine the most relevant node embeddings in relation to the text input, and consolidate these node embeddings into a holistic graph-level embedding. After that, GNP encompasses a domain projector to bridge the inherent disparities between the graph and text domains. Finally, a self-supervised link prediction objective is introduced to enhance the model comprehension of relationships between entities and capture graph knowledge in a self-supervised manner.

To fully evaluate our model, we conduct extensive experiments on multiple public benchmark datasets in the tasks of commonsense reasoning and biomedical reasoning. We further report the results across different LLM sizes and settings. We conclude that GNP can effectively encode intricate knowledge in KGs and significantly improve performance. Figure \ref{fig:intro_results} shows the averaged performance improvement using our method across six datasets. Specifically, GNP improves the baseline by \textbf{+13.5\%} when LLM is frozen, validating the superiority of our method in learning effective prompts. In addition, by using our method, fine-tuning LLMs with parameter-efficient approach LoRA \cite{lora} shows an improvement of \textbf{+1.8\%}. More promisingly, compared to model full fine-tuning without leveraging any efficient tuning approaches, our method can achieve competitive or superior performance in \textbf{10 out of 12} evaluations, as shown in the experiment section. To summarize, our main contributions are:

\begin{itemize}
    \item To the best of our knowledge, this is the first attempt to study the learning of beneficial knowledge from KGs for pre-trained LLMs.
    
    \item We propose GNP, a novel plug-and-play method for pre-trained LLMs to extract valuable knowledge from KGs. The proposed method contains various tailored designs, including a standard GNN, a cross-modality pooling module, a domain projector, and a self-supervised graph learning objective.
    
    \item Extensive experiments demonstrate the superiority of GNP on multiple datasets across different settings. 
    We also present the ablation study, model design comparison, parameter sensitivity analysis, case study and visualization to validate the effectiveness of GNP.

\end{itemize}

\begin{figure*}[t]
\centering
\includegraphics[width=\linewidth]{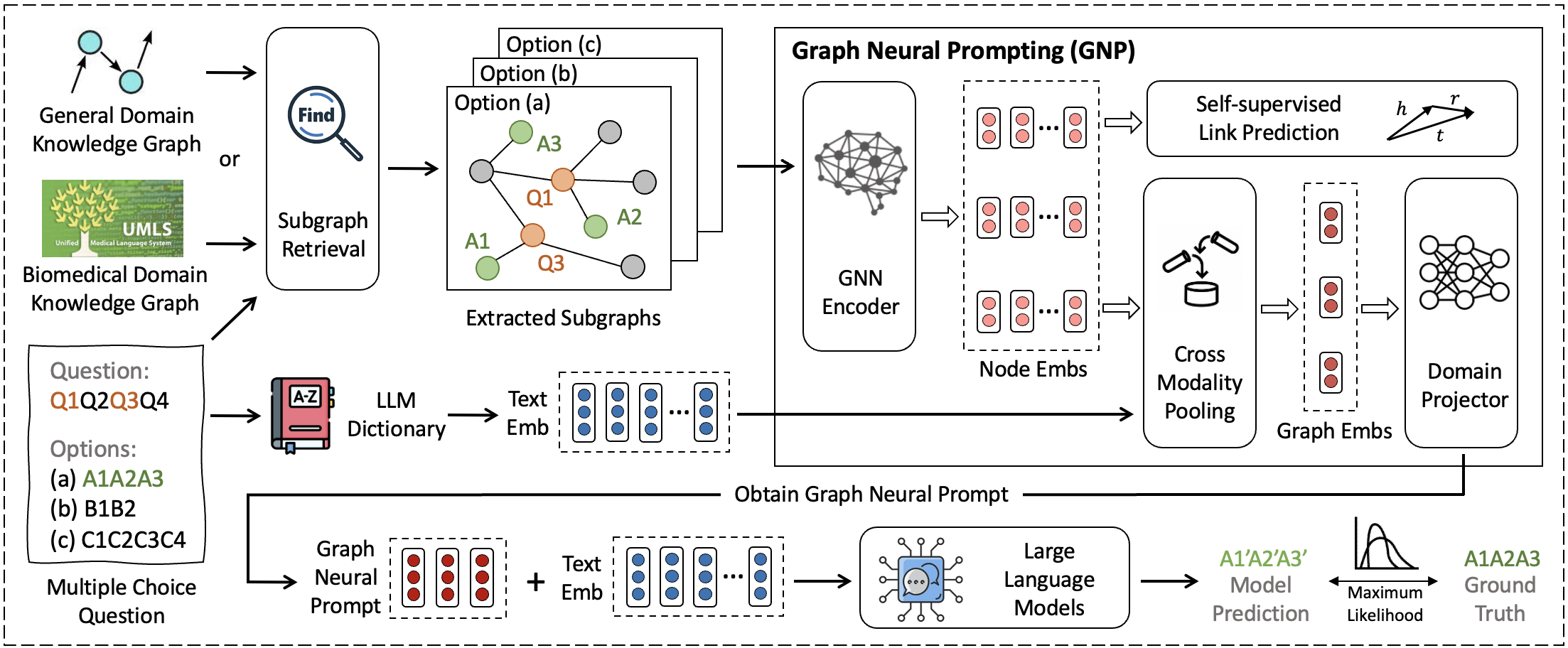}
\caption{
The overall framework. Given a multiple choice question, we first retrieve subgraphs from the knowledge graph based on the entities in the question and options. We then develop Graph Neural Prompting (GNP) to encode the pertinent factual knowledge and structural information to obtain the Graph Neural Prompt. GNP contains various designs including a GNN, a cross-modality pooling module, a domain projector, and a self-supervised link prediction objective. Later, the obtained Graph Neural Prompt is sent into LLM for inference along with the input text embedding. We utilize the standard maximum likelihood objective for downstream task adaptation, while LLM is kept frozen or tuned depending on different experimental settings. 
}
\label{fig:pipeline}
\end{figure*}

\section{Related Work}

\noindent
\textbf{Large Language Models and Question Answering.}
Recently, various LLMs have been proposed \cite{chung2022scaling, touvron2023llama, prompt_design} and have demonstrated remarkable performance across different tasks \cite{shi2023mededit, chen2023reason, wei2023llmrec, hong2023dp}. Question answering, as a fundamental task, demands intricate reasoning and understanding comprehension skills to interpret the text and provide appropriate responses to the posed questions \cite{lu2022learn, zhu2021retrieving, wang2023knowledge, chen2023minprompt}. Although LLMs have strong learning capabilities, they have the limitation of precisely capturing accurate factual knowledge and are susceptible to generating unfounded responses \cite{llm_survey, ji2023survey, bang2023multitask}. In addition, the enormous number of parameters in LLMs poses difficulties in adapting LLMs for downstream tasks \cite{scao2022bloom, smith2022using}. Correspondingly, various approaches are presented to alleviate the intensive training dependency and reduce the computational expenses \cite{prompt_tuning, prefix_tuning, lora}. For instance, Prompt Tuning \cite{prompt_tuning} introduces soft prompts to condition the pre-trained LLMs for downstream tasks. In our work, we propose to retrieve the factual knowledge from KGs to enhance LLMs, while still benefiting from circumventing the burdensome training expenses by using pre-trained LLMs.

\noindent
\textbf{Knowledge Graphs for Language Modeling.}
Many graph learning methods are proposed to encode graphs and KGs \cite{kg_survey, hgmae, nosmog, tang2022positive, wang2022graph, xu2023reusable, crowdgraph}. Recent studies indicate that KGs can enhance language modeling by providing background knowledge \cite{ren2021lego, wang2019improving}. One approach to achieve this is integrating KGs into the pre-training stage of language modeling. For instance, ERNIE \citep{sun2021ernie}, JAKET \citep{yu2022jaket}, and JointGT \citep{ke2021jointgt} develop pre-training objectives tailored for KG triples and the paired sentences. DRAGON \citep{dragon} introduces a customized fusion framework to jointly pre-train the model for KGs and text. Moreover, KGs are leveraged to assist language modeling for question answering \cite{kagnet, lv2020graph, feng2020scalable, mihaylov2018knowledgeable}. Specifically, GreaseLM \cite{greaselm} and QAGNN \cite{qagnn} suggest that KGs can scaffold reasoning about entities with the graph structure such as negation and multi-hop reasoning to facilitate complex question answering. To encode KGs, many works study methods to learn KG entity and relation embeddings, such as TransE \cite{transe} and DistMult \cite{distmult}. Recently, with the aim of integrating KGs into the emerging domain of LLMs, given existing studies pose difficulties when applying, KAPING \cite{kaping} employs knowledge graphs to extract relevant triples. These triples correspond to the input question, with the expectation that directly feeding them into LLMs is beneficial, despite the presence of noise. In our work, we present a learning method for identifying beneficial knowledge from KGs, offering substantial benefits to LLMs.

\section{Preliminary}
In this section, we describe the knowledge graph and formally define the problem of multiple choice question answering.

\begin{definition}
{\bf Knowledge Graph.} 
A knowledge graph is defined as $\mathcal{G}=(\mathcal{E}, \mathcal{R}, \mathcal{T})$, where $\mathcal{E}$ is the set of entities and $\mathcal{R}$ is the set of relations.
$\mathcal{T}$ is the collection of fact triples $\{(e_h, r, e_t)\} \in \mathcal{E} \times \mathcal{R} \times \mathcal{E}$, where $e_h$ denotes the head entity, $r$ is the relation, and $e_t$ indicates the tail entity.
\end{definition}

\begin{problem}
{\bf Multiple Choice Question Answering.}
Given a question $Q$, a set of answer options $A=\{a_k\}_{k=1}^K$, and an optional context $C$ depending on open-book or close-book, the task is to design a machine learning model $\mathcal{F}_\Theta$ with parameters $\Theta$ that selects the best option to answer the question. Here $K$ denotes the total number of answer options and $a_k$ indicates the $k$-th answer option. The ground truth label $y \in A$ is the correct answer for $Q$. In addition, we use knowledge graph $\mathcal{G}$ to provide rich knowledge and assist the model to answer the question. 
\end{problem}

\section{Methodology}

In this section, we introduce the techniques of prompting LLMs for question answering as well as subgraph retrieval. Additionally, we present Graph Neural Prompting and elaborate on its components and designs. Figure \ref{fig:pipeline} illustrates the framework of our method.

\subsection{Prompting LLMs for Question Answering}
Prompting is the de facto approach to elicit responses from LLMs \cite{prompt_survey}. The typical approach of prompting LLMs for multi-choice question answering is simple. Given a question $Q$, the optional context $C$, and the answer options $A$, we first tokenize the concatenation of $C, Q, A$ into a sequence of input text tokens $X$. We then design a series of prompt tokens, $P$, and prepend it to the input text tokens $X$, which is later considered as input for the LLM model to generate prediction $y' = f([P, X])$. The LLM model can be trained for downstream task adaptation using a standard maximum likelihood loss using teacher forcing \cite{teacher_forcing} and a cross-entropy loss:
\begin{equation}
    \mathcal{L}_{llm} = -\log p(y|X, \Theta),
    \label{eq:L_llm}
\end{equation}
where $p$ is the probability distribution parameterized by the model.
The prompt $P$ can be either a hard prompt in the form of textual input, or a soft prompt in the form of learnable embedding vectors.

Unlike existing methods that solely use a text string as the hard prompt, our Graph Neural Prompting approach encodes structural and factual information contained in the knowledge graph $\mathcal{G}$ into a soft prompt $P$, which is a sequence of trainable vectors that can be concatenated with the token embedding of $X$.  The learning of $P$ is encouraged to provide rich structural information and knowledge from $\mathcal{G}$ as well as task instruction for each data instance.

\subsection{Subgraph Retrieval}
To semantically align the input text tokens $X$ with the massive knowledge graph $\mathcal{G}$ with millions of nodes, we retrieve subgraphs of $\mathcal{G}$ that contain the relevant entities to the tokens in $X$. In particular, for each answer option $a_k$ and its corresponding context $C$ and question $Q$, we first obtain a set of matched entities $\mathcal{E}_{match}$ via entity linking to match the tokens in $X$ to the entities in $\mathcal{G}$. We then retrieve a subgraph $\mathcal{G}'$ based on the entities in $\mathcal{E}_{match}$ by including their two-hop neighbors and the relations that connect them \cite{dragon}. The retrieved subgraph contains the necessary content and knowledge to assist the model in answering $Q$.

\subsection{Graph Neural Prompting}

Graph Neural Prompting contains various designs, including a GNN encoder that embeds the knowledge graph, a cross-modality pooling module that determines the pertinent node embeddings, a domain projector that bridges the discrepancies between graph and text, and a self-supervised link prediction objective that encourages the model to recognize structural information.

\noindent
\textbf{GNN Encoder.}
Although the retrieved subgraph $\mathcal{G}'$ contains rich contextual information regarding the question and answer choices, some entities and relations are not relevant to the actual question. Directly feeding every fact triples in $\mathcal{G}'$ can introduce noise and prevent the LLM model from concentrating on the critical information. Therefore, we introduce a GNN to encode the most relevant knowledge and further integrate the complex relationships among the entities. In particular, we first initialize the node embeddings using pre-trained entity embeddings \cite{feng2020scalable, linkbert}. Next, we employ a standard graph attention network \cite{gat} as our GNN encoder for the retrieved subgraph $\mathcal{G}'$. The encoding process is formulated as follows:
\begin{equation}
    H_1 = f_{GNN}(\mathcal{G}'),
\end{equation}
where $H_1 \in \mathbb{R}^{d_g}$ represents the node embeddings learned by GNN for every node in $\mathcal{G}'$, and $d_g$ denotes the output dimension of the GNN encoder.

\noindent
\textbf{Cross-modality Pooling.}
With the aim of identifying the most pertinent nodes in relation to the question, and consolidating the node embeddings into a holistic graph-level representation for subsequent use, we design the cross-modality pooling module. In particular, we first introduce a self-attention layer to dynamically identify node significance using the internal graph characteristics and the implicit interactions among nodes:
\begin{equation}
    H_2 = \text{Self-Attn}(H_1),
\end{equation}
where $H_2$ is node embeddings obtained after calculating self-attention and $\text{Self-Attn}$ indicates the self-attention component. Then, we leverage the textual prompt to calculate the importance of nodes within the graph. To ensure uniformity, we utilize the dictionary in the LLM to obtain the text embeddings $\mathcal{T} \in \mathbb{R}^{d_t}$ for every token in the input text, where $d_t$ denotes the dimension of the LLM dictionary. Concretely, we start by applying a transformation to the text embeddings $\mathcal{T}$ and obtain the transformed text embedding $\mathcal{T}'$, ensuring that the dimension of $\mathcal{T}'$ matches the dimension $d_g$ of node embeddings $H_2$. After that, we calculate the cross-modality attention using $H_2$ and $\mathcal{T}'$. We use $H_2$ as the query and the $\mathcal{T}'$ as the key and the value. The procedure is as follows:
\begin{equation}
\begin{aligned}
    \mathcal{T}' &= \text{FFN}_1(\sigma (\text{FFN}_2(\mathcal{T}))), \\
    H_3 &= \text{softmax}[H_2 \cdot (\mathcal{T}')^T / \sqrt{d_g}] \cdot \mathcal{T}',
\end{aligned}
\end{equation}
where $\sigma$ is the GELU activation function, $\text{FFN}_1$ and $\text{FFN}_2$ are feed-forward neural networks, and $H_3$ is the final node embeddings obtained with cross-modality attention considered. Next, we generate the graph-level embedding by average pooling the node embeddings $H_3$ in $\mathcal{G}'$: 
\begin{equation}
    H_4 = \text{POOL}(H_3),
\end{equation}
where $H_4$ represents the graph-level embedding that takes into account the node significance in $\mathcal{G}'$.

\noindent
\textbf{Domain Projector.} 
In order to create a mapping between the graph-level embeddings and the text domain to facilitate comprehension by the LLM, we design a domain projector to align them. This projector aims to bridge the inherent disparities between the graph and text, allowing for more seamless integration. In addition, the projector maps the graph-level embeddings to the same dimension $d_t$ of LLM, which ensures compatibility and consistency when interfacing with the LLM's inherent structures. We design the projector as follows:
\begin{equation}
    Z = \text{FFN}_3(\sigma (\text{FFN}_4(H_4))), 
\end{equation}
where $Z$ denotes Graph Neural Prompt, the final output of GNP, and $\text{FFN}_3, \text{FFN}_4$ are feed-forward neural networks.

\noindent
\textbf{Self-supervised Link Prediction.} 
While the downstream cross-entropy objective enables the model to learn and adapt to the target dataset, we design a link prediction task to further refine its understanding of relationships between entities and capture graph knowledge in a self-supervised manner. Specifically, we mask out some edges in $\mathcal{G}'$ and enforce the model to predict them. This encourages the model to learn to use the partial graph content and structure to reason about the missing links. Concretely, we denote the set of masked-out edges as $\mathcal{E}_{mask} \subseteq \mathcal{E}$. Given the learned node embeddings of the head entity and tail entity in a triplet $\{\mathbf{h}_3, \mathbf{t}_3\} \in H_3$, we adopt a widely-used knowledge graph embedding method DistMult \cite{distmult} to map the entity embeddings and relation in the KG to vectors, $\mathbf{h}, \mathbf{r}, \mathbf{t}$. We then define the scoring function $\phi(e_h, e_t) = \langle\mathbf{h}, \mathbf{r}, \mathbf{t} \rangle$ to generate the scores for each triple, where $\langle \cdot, \cdot, \cdot \rangle$ denotes the trilinear dot product, and $\mathbf{r}$ represents the relations in KGs. A higher $\phi$ indicates a higher chance of $(e_h,r,e_t)$ being a correct positive triple instead of an incorrect negative triple. We enforce the model to predict the masked edges in $\mathcal{E}_{mask}$ as positive and other random edges as negative. The link prediction loss $\mathcal{L}_{lp}$ is defined as follows:
\begin{equation}
\mathcal{L}_{lp} = \sum_{(e_h,r,e_t)\in {\mathcal{E}_{mask}}} (S_{pos} + S_{neg}),
\end{equation}
where $S_{pos} = -\log{\sigma_s(\phi(e_h, e_t) +\gamma)}$ indicates the score for correct positive triples, $\gamma$ is the margin, $\sigma_s$ is the sigmoid function, $\{(e_h',r,e_t')\}$ are $n$ negative triples corresponding to the positive triplet $(e_h, r, e_t)$, and $S_{neg} = \frac{1}{n}\sum_{(e_h', r, e_t')}\log{\sigma_s(\phi(e_h', e_t') +\gamma)}$ is the score for incorrect negative triples. The final objective function $\mathcal{L}$ is defined as the weighted combination of $\mathcal{L}_{llm}$ and $\mathcal{L}_{lp}$: 
\begin{equation}
\mathcal{L} = \mathcal{L}_{llm} + \lambda \mathcal{L}_{lp},
\end{equation}
where $\lambda$ is a trade-off weight for balancing two losses.

\begin{table*}[t]
\centering
\caption{Overall experimental results on commonsense reasoning and biomedical reasoning tasks. The best results across different LLM sizes and settings are highlighted in bold. $\Delta_{PT}$ and $\Delta_{LoRA}$ represent the relative performance improvement of our method to Prompt Tuning and LoRA, respectively. We also include the full fine-tuning result in gray color for further reference. * means multiple prompt design methods are evaluated while only the best result is reported. Accuracy is used as the evaluation metric.
}
\label{tab:main-results}
\resizebox{\linewidth}{!}{
\begin{NiceTabular}{cccccccccc}
\toprule
 &  &  & \multicolumn{4}{c}{Commonsense Reasoning} & \multicolumn{2}{c}{Biomedical Reasoning} \\ 
 \cmidrule{4-9}
\multirow{-2.3}{*}{\textbf{LLM}} & \multirow{-2.3}{*}{\textbf{Setting}} & \multirow{-2.3}{*}{\textbf{Method}} & \textbf{OBQA} & \textbf{ARC} & \textbf{PIQA} & \textbf{Riddle} & \textbf{PubMedQA} & \textbf{BioASQ} & \multirow{-2.3}{*}{\textbf{Total}} \\ 

\midrule

\multirow{13}{*}{\textbf{\begin{tabular}[c]{@{}c@{}}FLAN-T5 \\ xlarge (3B)\end{tabular}}}
 & \multirow{9}{*}{\begin{tabular}[c]{@{}c@{}}LLM Frozen\end{tabular}} 
 
 & LLM-only & 69.20 & 68.24 & 58.43 & 53.73 & 71.50 & 65.85 & 64.49 \\
 &  & Prompt Designs* & 72.20 & 70.99 & 60.94 & 52.75 & 70.50 & 67.48 & 65.33 \\
 &  & KG Flattening REL & 61.80 & 64.12 & 57.56 & 43.33 & 69.25 & 65.04 & 60.18 \\
 &  & KG Flattening BFS & 62.80 & 63.86 & 56.69 & 44.12 & 69.25 & 65.04 & 60.29 \\
 &  & KAPING TH & 58.80 & 63.52 & 52.34 & 40.78 & 70.00 & 65.04 & 58.41 \\
 &  & KAPING OH & 60.00 & 63.09 & 51.69 & 41.37 & 70.00 & 65.04 & 58.53 \\
 &  & Prompt Tuning & 72.20 & 70.64 & 60.83 & 53.33 & 72.00 & 66.67 & 65.95 \\
 &  & \frameworkabbr & \textbf{79.80} & \textbf{71.85} & \textbf{61.48} & \textbf{66.86} & \textbf{76.75} & \textbf{89.43} & \textbf{74.36} \\
 &  & $\Delta_{PT}$ & $\uparrow$ 10.53\% & $\uparrow$ 1.71\% & $\uparrow$ 1.07\% & $\uparrow$ 25.37\% & $\uparrow$ 6.60\% & $\uparrow$ 34.14\% & $\uparrow$ 12.76\% \\
 \cmidrule{2-10}
 
 & \multirow{4}{*}{LLM Tuned} & Full Fine-tuning & {\color[HTML]{9B9B9B} 82.80} & {\color[HTML]{9B9B9B} 73.30} & {\color[HTML]{9B9B9B} 63.55} & {\color[HTML]{9B9B9B} 74.12} & {\color[HTML]{9B9B9B} 76.25} & {\color[HTML]{9B9B9B} 91.06} & {\color[HTML]{9B9B9B} 76.85} \\
 & & LoRA & 80.40 & 71.33 & 63.76 & 72.94 & 76.25 & 92.68 & 76.23 \\
 &  & LoRA + \frameworkabbr & \textbf{83.40} & \textbf{72.45} & \textbf{64.31} & \textbf{75.49} & 76.25 & 92.68 & \textbf{77.43} \\
 &  & $\Delta_{LoRA}$ & $\uparrow$ 3.73\% & $\uparrow$ 1.57\% & $\uparrow$ 0.86\% & $\uparrow$ 3.50\% & $\uparrow$ 0.00\% & $\uparrow$ 0.00\% & $\uparrow$ 1.58\% \\

\midrule
\multirow{13}{*}{\textbf{\begin{tabular}[c]{@{}c@{}}FLAN-T5 \\ xxlarge (11B)\end{tabular}}}
 & \multirow{9}{*}{\begin{tabular}[c]{@{}c@{}}LLM Frozen\end{tabular}} 
 & LLM-only & 76.80 & 68.93 & 56.58 & 61.37 & 71.75 & 65.85 & 66.88 \\
 &  & Prompt Designs* & 79.60 & 74.16 & 58.00 & 60.59 & 71.25 & 66.67 & 68.38 \\
 &  & KG Flattening REL & 72.80 & 66.78 & 56.80 & 53.53 & 69.50 & 66.67 & 64.35 \\
 &  & KG Flattening BFS & 72.40 & 66.95 & 56.37 & 54.90 & 68.75 & 65.85 & 64.20 \\
 &  & KAPING TH & 60.60 & 57.25 & 53.21 & 48.43 & 68.75 & 66.67 & 59.15 \\
 &  & KAPING OH & 60.00 & 56.65 & 52.99 & 47.65 & 69.25 & 66.67 & 58.87 \\
 &  & Prompt Tuning & 78.80 & 74.85 & 61.26 & 61.37 & 70.00 & 65.04 & 68.55 \\
 &  & \frameworkabbr & \textbf{87.20} & \textbf{78.20} & \textbf{63.66} & \textbf{70.98} & \textbf{76.75} & \textbf{90.24} & \textbf{77.84} \\
 &  & $\Delta_{PT}$ & $\uparrow$ 10.66\% & $\uparrow$ 4.48\% & $\uparrow$ 3.92\% & $\uparrow$ 15.66\% & $\uparrow$ 9.64\% & $\uparrow$ 38.75\% & $\uparrow$ 13.54\% \\
 
 \cmidrule{2-10}
  & \multirow{4}{*}{LLM Tuned} & Full Fine-tuning & {\color[HTML]{9B9B9B} 89.40} & {\color[HTML]{9B9B9B} 76.82} & {\color[HTML]{9B9B9B} 65.61} & {\color[HTML]{9B9B9B} 80.78} & {\color[HTML]{9B9B9B} 78.00} & {\color[HTML]{9B9B9B} 92.68} & {\color[HTML]{9B9B9B} 80.55} \\
 & & LoRA & 88.60 & 78.54 & 65.61 & 74.90 & 77.75 & 91.06 & 79.41 \\
 &  & LoRA + \frameworkabbr & \textbf{89.60} & \textbf{78.71} & \textbf{65.94} & \textbf{76.67} & \textbf{79.75} & \textbf{94.31} & \textbf{80.83} \\
 &  & $\Delta_{LoRA}$ & $\uparrow$ 1.13\% & $\uparrow$ 0.22\% & $\uparrow$ 0.50\% & $\uparrow$ 2.36\% & $\uparrow$ 2.57\% & $\uparrow$ 3.57\% & $\uparrow$ 1.79\% \\

\bottomrule

\end{NiceTabular}
}
\end{table*}

\section{Experiments}

In this section, we conduct extensive experiments to compare the performances of different models. We also show ablation study, model design comparison, and parameter sensitivity analysis to demonstrate the effectiveness of GNP. Moreover, we present case study and visualization to provide an intuitive understanding and illustrate how KGs benefit.

\subsection{Experiment setup}

\textbf{Knowledge Graphs and Datasets.}
We conduct experiments on both the general domain (commonsense reasoning) and the biomedical domain (biomedical reasoning). For the used knowledge graphs, we consider ConceptNet \cite{conceptnet} that contains rich commonsense knowledge regarding the daily concepts, and Unified Medical Language System (UMLS) \cite{umls} that involves well-structured health and biomedical information. For datasets, we use four commonsense reasoning datasets, including OpenBookQA (OBQA) \cite{mihaylov2018can}, AI2 Reasoning Challenge (ARC) \cite{clark2018think}, Physical Interaction Question Answering (PIQA) \cite{bisk2020piqa}, and RiddleSense (Riddle) \cite{lin-etal-2021-riddlesense}. In addition, we consider PubMedQA \cite{jin2019pubmedqa} and BioASQ \cite{tsatsaronis2015overview} for biomedical reasoning.

\noindent
\textbf{Two Settings: LLM Frozen vs. LLM Tuned.}
To fully evaluate the model, we employ two settings: LLM Frozen and LLM Tuned. For LLM Frozen, we keep the parameters in LLM unchanged and only adapt the prompt. For LLM Tuned, the original LLM parameters are updated for downstream tasks by utilizing LoRA or full fine-tuning.

\noindent
\textbf{Baselines.}
In the setting of LLM Frozen, we compare with nine baselines, including LLM-only that uses no prompt, three prompt design methods that use different instructions as hard prompts, KG Flattening that flattens the nodes in the graph into a sequence via relevance score (REL) ranking \cite{dragon} or breadth-first search (BFS), KAPING \cite{kaping} that injects the important KG triples within one-hop (OH) and two-hop (TH) neighborhoods, and Prompt Tuning \cite{prompt_tuning} that introduces soft prompts. In the setting of LLM Tuned, we compare with LoRA that updates partial LLM parameters. In addition, we include full model fine-tuning results as the referencing benchmark. 

\noindent
\textbf{Implementation Details.}
For the proposed model, we set the learning rate to 1e-4, batch size to 8, hidden dimension of GNN to 1024, and training epochs to 50. In order to adapt the model effectively to each dataset, we search the GNN layers from 2 to 5, cross-modality pooling layers from 1 to 3, trade-off weight $\lambda$ from \{0.1, 0.5\}, and link drop rate from \{0.1, 0.3, 0.7\}. We choose FLAN-T5 xlarge (3B parameters) and xxlarge (11B parameters) as the LLMs used in this paper.  We adjust the maximum sequence length of LLMs to best fit the question length for each dataset. We run all experiments on four NVIDIA Tesla V100 GPUs with 24GB RAM.

\subsection{Performance Comparison}
To comprehensively evaluate our model, we conduct rigorous experiments using various LLMs across two reasoning tasks under different settings. The results are reported in Table \ref{tab:main-results}. According to the table, in the setting of LLM Frozen, we observe that the utilization of the prompt design instructions often yields performance improvement, compared to LLM-only that uses no instructions, though the enhancement is mostly marginal. 
Interestingly, the baseline methods that inject KG information directly (KG Flattening and KAPING) can significantly hurt the model performance. This aligns with our motivation that KGs contain irrelevant contexts for the downstream tasks that could introduce noises or even alter the semantics if not handled carefully. 
While Prompt Tuning shows improved outcomes using the trainable soft prompts, their improvement is trivial. In contrast, our \frameworkabbr exhibits significant and notable performance improvements across various datasets, settings, and LLMs. For example, for the commonsense reasoning task, GNP provides +25.37\% improvement on Riddle for 3B LLM, and +15.66\% improvement for 11B LLM. In addition, for the biomedical reasoning task, GNP improves the performance by +34.14\% on BioASQ for 3B LLM and +38.75\% for 11B LLM. In general, GNP achieves an improvement of \textbf{+12.76\%} and \textbf{+13.54\%} for 3B and 11B LLM, respectively.

In the setting of LLM Tuned, we first study the performance in comparison with LoRA and then report the model full fine-tuning for additional reference. As shown in the table, LoRA is a significantly more powerful approach than Prompt Tuning due to the direct update of the LLM internal parameters. Combining with the proposed GNP, the performance can be further improved. For example, GNP achieves 3.73\% improvement on OBQA for 3B LLM, and 3.57\% improvement on BioAQS for 11B LLM. Moreover, model full fine-tuning is an important reference to study the performance gap since LoRA only updates a small fraction of the model parameters. Surprisingly, we find that the incorporation of GNP can surpass the results of full fine-tuning. In contrast, relying solely on LoRA shows difficulties in achieving a comparable performance of full fine-tuning. In total, our final performance matches or surpasses model full fine-tuning in \textbf{10 out of 12} evaluations across different LLM sizes and datasets, as shown in Table \ref{tab:main-results}.

\begin{table}[t]
\centering
\caption{
Results of ablation study.
}
\label{tab:ablation}
\resizebox{\columnwidth}{!}{
\begin{NiceTabular}{cccccc}
\toprule
 &  &  \multicolumn{2}{c}{Commonsense} & \multicolumn{2}{c}{Biomedical} \\ 
 \cmidrule{3-6}
\multirow{-2.3}{*}{\textbf{LLM}} & \multirow{-2.3}{*}{\textbf{Variant}}
& \textbf{OBQA} & \textbf{ARC} & \textbf{PubMedQA} & \textbf{BioASQ} \\ 
\midrule
\multirow{4}{*}{\textbf{\begin{tabular}[c]{@{}c@{}}FLAN-T5 \\ xlarge (3B)\end{tabular}}} 
 & w/o CMP & 78.00 & 69.44 & 76.00 & 86.18 \\
 & w/o SLP & 78.80 & 69.18 & 75.75 & 88.62 \\
 & w/o DP & 73.00 & 70.30 & 76.25 & 83.74 \\
 & GNP & \textbf{79.80} & \textbf{71.85} & \textbf{76.75} & \textbf{89.43} \\
\midrule

\multirow{4}{*}{\textbf{\begin{tabular}[c]{@{}c@{}}FLAN-T5 \\ xxlarge (11B)\end{tabular}}} 
 & w/o CMP & 85.20 & 76.91 & 75.75 & 87.80 \\
 & w/o SLP & 83.60 & 76.74 & 73.25 & 89.43 \\
 & w/o DP & 79.40 & 74.59 & 71.75 & 85.37 \\
 & GNP & \textbf{87.20} & \textbf{78.20} & \textbf{76.25} & \textbf{90.24} \\
 \bottomrule

\end{NiceTabular}
}
\end{table}

\subsection{Ablation Study}

Since GNP contains various model components (i.e., cross-modality pooling (CMP), self-supervised link prediction (SLP), and domain projector (DP)), we conduct ablation studies to analyze the contributions of different components by removing each of them independently (see Table \ref{tab:ablation}). Specifically, removing DP significantly affects the performance, showing that DP has a large contribution to the proposed method. In addition, the decreasing performances of removing CMP and SLP demonstrate the effectiveness of CMP and SLP in enhancing the model. In most cases, SLP yields greater significance compared to CMP, while in BioASQ, CMP plays a more important role. Finally, the proposed GNP achieves the best results in all cases, indicating the strong capability of different components in our model.

\subsection{Model Design Comparison}
A salient property of GNP is the learning of Graph Neural Prompt for each data instance, i.e., various questions yield different retrieved subgraphs, resulting in unique prompts. Given its distinction to the dataset-level prompt (DLP) from Prompt Tuning that learns prompt for each dataset, we present the outcomes of integrating DLP for further investigation. As shown in Table \ref{tab:design_integration}, incorporating DLP cannot further boost the performance and might even diminish it in certain cases. This indicates that our instance-level prompt provides adequate guidance for LLM to perform well. In addition, we validate the importance of explicitly modeling relations using a widely-used Relational GNN (RGNN) \cite{greaselm}. The observed decline in performance suggests that a standard GNN is sufficient to capture the graph information, and explicitly modeling the relations might increase the difficulty of generating suitable guidance for the task.

\begin{table}[t]
\centering
\caption{
Results of integrating different model designs.
}
\label{tab:design_integration}
\resizebox{\columnwidth}{!}{
\begin{NiceTabular}{cccccc}
\toprule
 &  &  \multicolumn{2}{c}{Commonsense} & \multicolumn{2}{c}{Biomedical} \\ 
 \cmidrule{3-6}
\multirow{-2.3}{*}{\textbf{LLM}} & \multirow{-2.3}{*}{\textbf{Design}}
& \textbf{OBQA} & \textbf{ARC} & \textbf{PubMedQA} & \textbf{BioASQ} \\ 
\midrule
\multirow{3}{*}{\textbf{\begin{tabular}[c]{@{}c@{}}FLAN-T5 \\ xlarge (3B)\end{tabular}}} 
 & GNP & \textbf{79.80} & \textbf{71.85} & \textbf{76.75} & \textbf{89.43} \\
 & + DLP & \textbf{79.80} & 70.30 & 75.50 & \textbf{89.43} \\
 & + RGNN & 79.00 & 71.49 & 75.50 & \textbf{89.43} \\ 
\midrule

\multirow{3}{*}{\textbf{\begin{tabular}[c]{@{}c@{}}FLAN-T5 \\ xxlarge (11B)\end{tabular}}} 
 & GNP & \textbf{87.20} & \textbf{78.20} & \textbf{76.25} & \textbf{90.24} \\
 & + DLP & 86.20 & 76.05 & 75.00 & 88.62 \\
 & + RGNN & 85.20 & 76.48 & 75.25 & 89.43 \\ 
 \bottomrule

\end{NiceTabular}
}
\end{table}

\subsection{Parameter Sensitivity}

Next, we perform sensitivity analysis focusing on the following parameters: the number of GNN layers and the number of layers in the cross-modality pooling component.

\begin{figure}[t]
	\centering
	\includegraphics[width=\columnwidth]{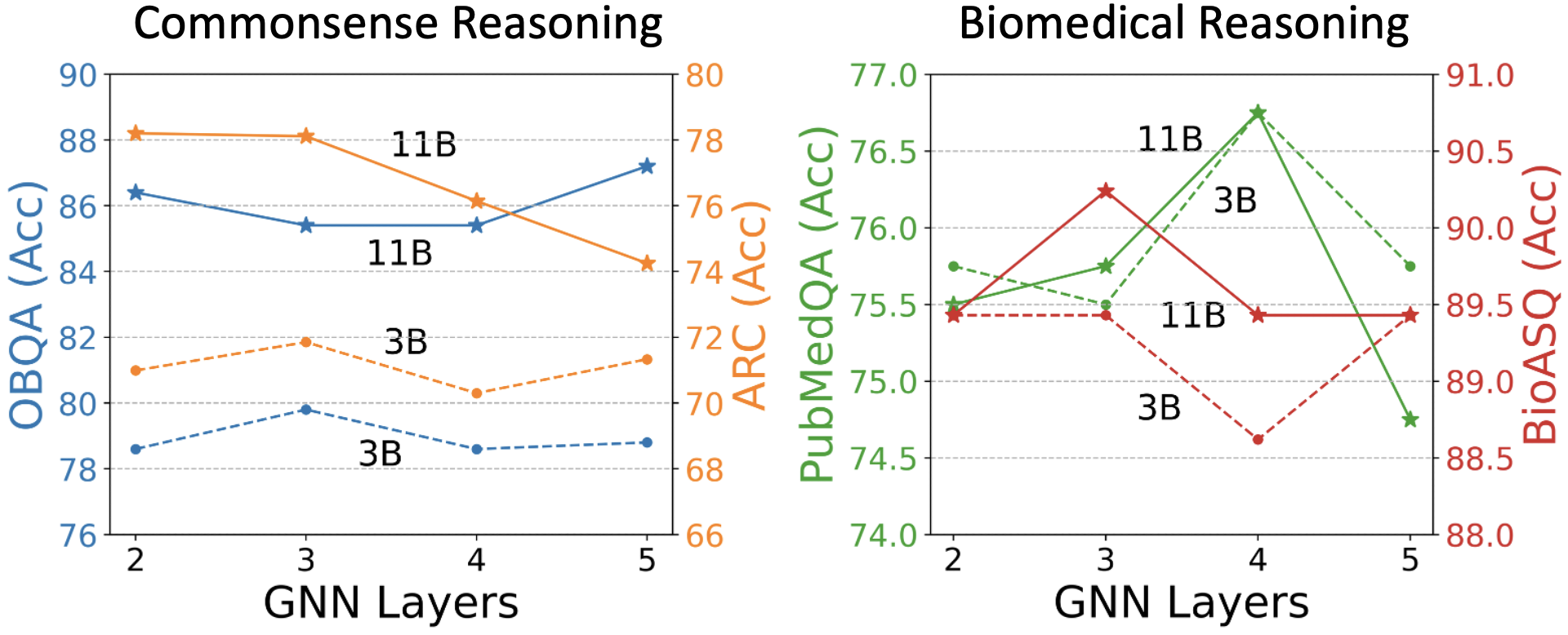}
	\caption{Performance \textit{w.r.t.} different number of GNN layers.}
	\label{fig:param_sens_gnn_layers}
\end{figure}

\noindent
\textbf{Impact of GNN layers.}
We evaluate the influence of GNN layers for both 3B and 11B models in Figure \ref{fig:param_sens_gnn_layers}. According to the figure, we have the following observations. First, various datasets have different optimal numbers of GNN layers. To illustrate, for ARC, 3 layers can achieve the optimal performance while 4 layers perform the best for PubMedQA. Second, the optimal number of GNN layers for 3B and 11B LLMs differs. For example, for OBQA, 3 layers work best for 3B LLM, while 11B LLM reaches its top performance when using 5 layers. Third, choosing different GNN layers can have a weak impact on some datasets while can also drastically affect the performance on other datasets. To demonstrate, increasing from 3 layers to 5 layers for 11B LLM can decrease the performance on ARC by a large margin (from 78.1 to 74.3), while adjusting the layers for BioASQ may not lead to a big change in the performance.

\begin{figure}[t]
	\centering
	\includegraphics[width=\columnwidth]{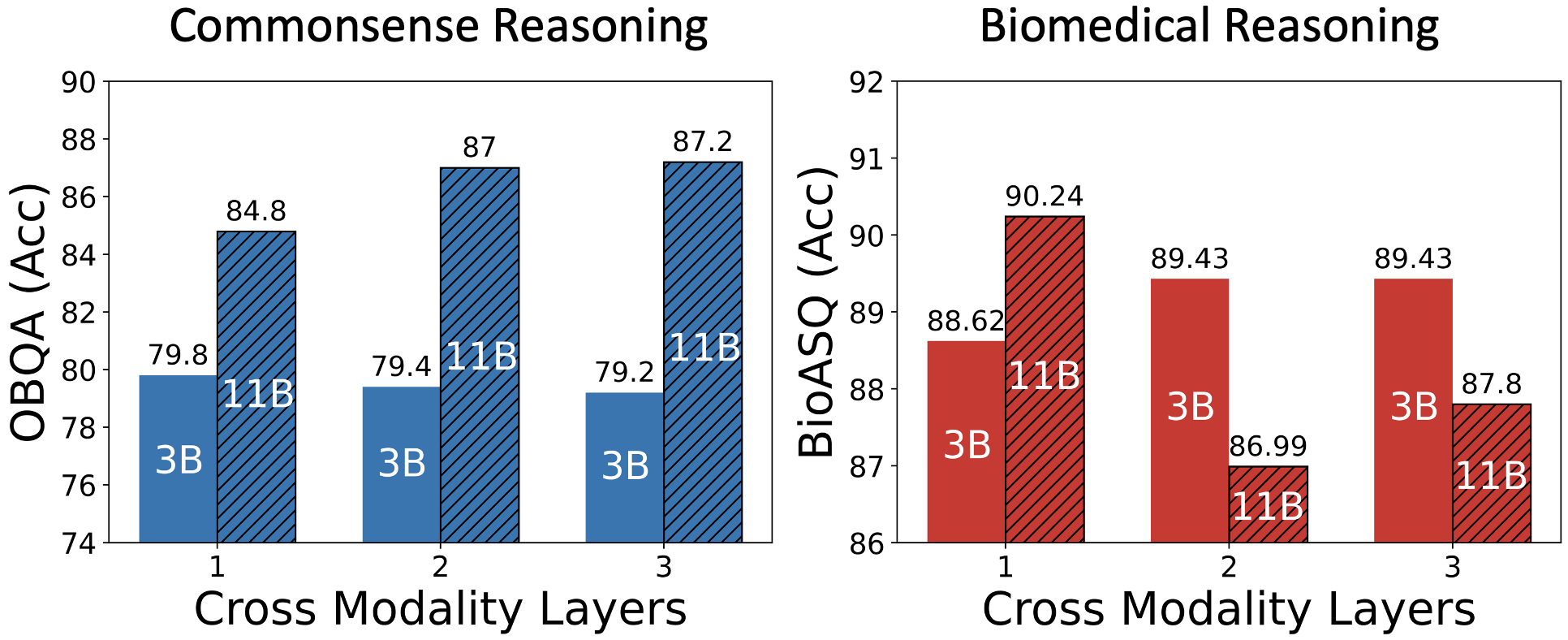}
	\caption{Performance \textit{w.r.t.} different number of cross-modality pooling layers.}
	\label{fig:param_sens_cross_mod_layers}
\end{figure}

\noindent
\textbf{Impact of cross-modality pooling layers.}
We report the performance of different cross-modality pooling layers in Figure \ref{fig:param_sens_cross_mod_layers}. As shown in the figure, we observe that the commonsense reasoning dataset OBQA and biomedical reasoning dataset BioASQ demonstrate different reactions to layer numbers. Specifically, for OBQA, the performance of the larger 11B LLM increases with more layers, while the performance of the smaller 3B LLM decreases. On the other hand, for BioASQ, the larger 11B LLM tends to show a degraded performance when adding more layers, while the smaller 3B model presents an improved performance. This indicates that suitable cross-modality pooling layers can lead to the best model performance.

\begin{figure}[ht]
	\centering
	\includegraphics[width=\linewidth]{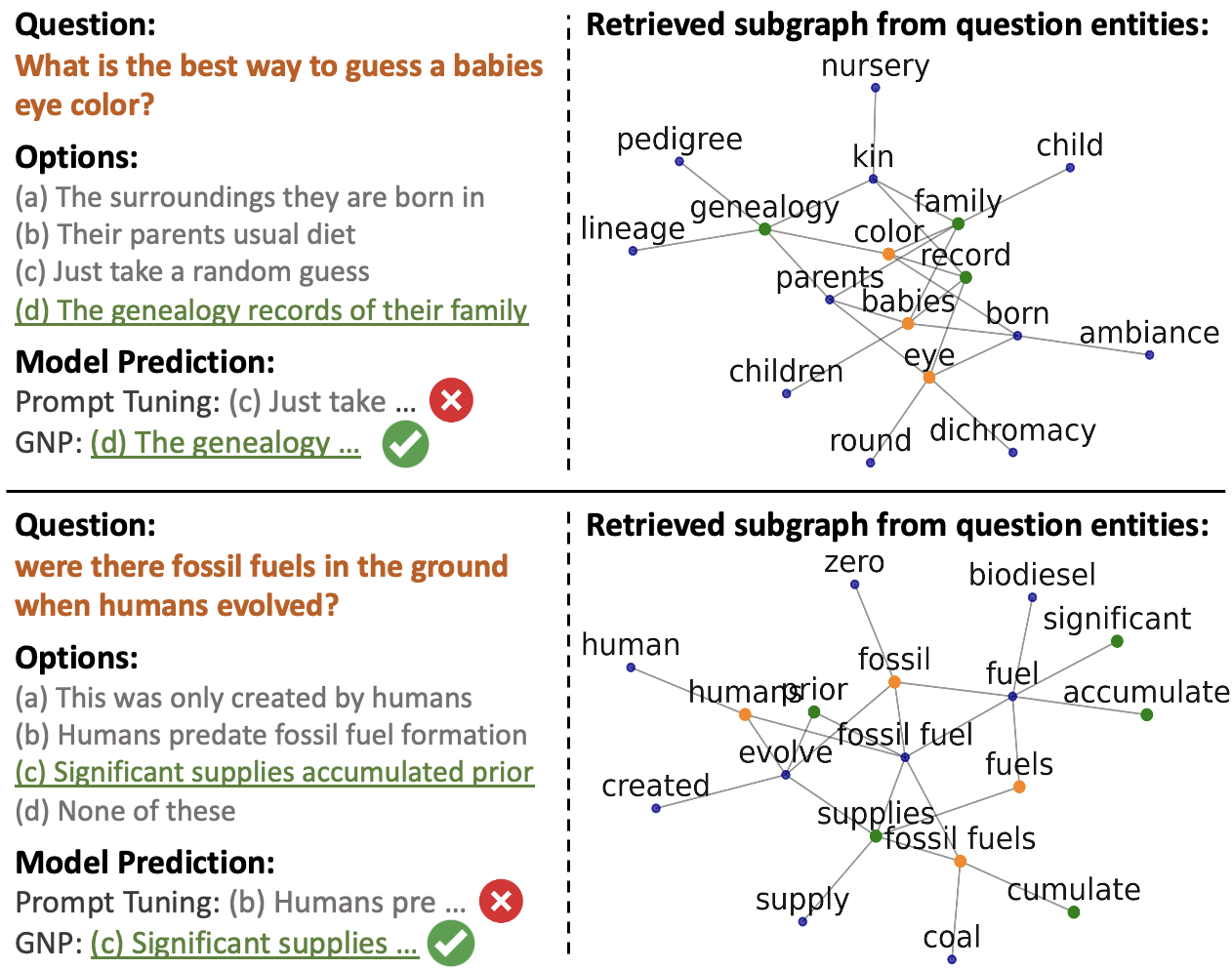}
	\caption{Case study on two QA examples from OBQA dataset. Question entities are marked in green and their subsampled neighbors in the KG are marked in blue. The entities appearing in the correct answer are marked in orange.
	}
	\label{fig:case_study}
\end{figure}

\subsection{Case Study and Visualization}

For a more intuitive understanding and comparison, we randomly select two examples from the OBQA dataset and visualize the retrieved subgraphs in Figure \ref{fig:case_study}. For visualization clarity, we only show question entities and a limited number of their neighbors. We remarkably notice that the retrieved subgraphs encompass certain entities for the correct answer, and there exist edges connecting the question and answer entities, which makes the task of question answering easier by leveraging this information.

To answer the question ``What is the best way to guess a babies eye color?", Prompt Tuning makes the wrong generation ``Just take a random guess". On the other hand, our retrieved subgraph offers the links that directly relate the entity ``babies" to ``family", ``record", and further to ``genealogy", which all appear in the correct option (d). This important context provides valuable insights for the model. Note that the subgraph also contains irrelevant entities such as ``round" and ``nursery". This explains why directly using the knowledge graph can introduce noise. However, our \frameworkabbr method possesses the capability to collect the most critical information in the graph to determine the correct answer.

The second question ``were there fossil fuels in the ground when humans evolved?" requires correctly identifying the historical sequencing order between the entity ``humans" and ``fossil fuels". The retrieved subgraph contains the critical relation, i.e., ``humans", ``evolve", ``prior", ``fossil fuel". Nevertheless, the subgraph also contains the entity ``created" that could confuse the model into selecting option (a). \frameworkabbr is able to capture the structural proximity among the key entities and select the correct answer (c).

\section{Conclusion}

In this paper, we address the limitations of LLMs in precisely capturing and returning grounded knowledge. In particular, we propose Graph Neural Prompting (GNP), a novel plug-and-play method to assist pre-trained LLMs in learning beneficial knowledge from KGs. Extensive experiments on commonsense and biomedical reasoning tasks demonstrate that GNP can improve the performance by \textbf{+13.5\%} when LLM is frozen, and \textbf{+1.8\%} when LLM is tuned. In addition, we present ablation studies, model design comparison, parameter sensitivity, case study and visualization to validate the effectiveness of the proposed method.

\bibliography{aaai24}

\end{document}